%% file: submission.tex
\documentclass[10pt]{article}

\usepackage{wsstyle}

\usepackage[utf8]{inputenc}
\usepackage[T1]{fontenc}
\usepackage{mathptmx}
\usepackage[scaled=0.92]{helvet}
\usepackage{courier}
\usepackage{microtype}
\usepackage{amsmath,amssymb}
\usepackage{graphicx}
\usepackage{booktabs}
\usepackage{xcolor}
\usepackage[numbers,sort&compress]{natbib}
\usepackage[colorlinks=true,allcolors=blue,breaklinks=true]{hyperref}
\usepackage{url}

\newcommand{\vhat}{\hat{v}}
\newcommand{\kap}{\kappa}

\newcommand{\inputtable}[1]{%
  \IfFileExists{tables/#1.tex}{\input{tables/#1}}{%
    \begin{table}[t]\centering\small
    \fbox{\parbox{0.92\linewidth}{\textbf{MISSING:
    \texttt{\detokenize{tables/}\detokenize{#1}\detokenize{.tex}}.}
    Generated by \texttt{analysis/make\_tables.py} from a run manifest; this
    build had no such run. \emph{Do not submit this PDF.}}}
    \end{table}}}

\title{Cosine Similarity Is Not Evidence:\\
{\large Measuring the Noise Floor of Interpretability Transfer Under Quantization}}

\author{%
  Pranav Varshney\\
  University of Michigan\\
  \texttt{pvarsh@umich.edu}%
}
\date{}

\begin{document}
\maketitle

\begin{abstract}
A statistic reported without the quantity needed to interpret it is not
evidence. We develop that thesis for a concrete practice in AI safety.
Interpretability artifacts are calibrated on full-precision weights, deployed
on quantized ones, and certified as surviving the change by scale-invariant
statistics (cosine similarity, correlation, AUROC) that are reported without
their noise floor. For the difference-in-means direction estimator, the
split-half floor is governed by one dimensionless number, $\kap = n\rho^2/d$.
The closed form $\mathbb{E}[\cos] \approx (1+4/\kap)^{-1}$ is classical; the
missing input is the class separation $\rho$, which we measure on real
activations; no compression-transfer study we know of reports it. On
Qwen2.5-1.5B-Instruct, $\rho = 33$--$61$ across depth, so two independent runs
of the estimator agree to $0.978$--$0.994$ by sampling alone. A published
cosine of $0.996$ between full-precision and quantized refusal directions
therefore cannot be read as preservation without the $n$ it was computed at,
which is not reported. Where $n$ is known, we judge each low-bit cosine against
the split-half null measured \emph{within that quantized model}, because a
full-precision null assumes the low-bit estimator has the same variance. That
assumption is exactly what a null exists to test. The result is plain: at INT4
the direction rotated, and the deficit exceeds the estimator's own noise. At
INT8 we detect no movement, which is not an equivalence claim. We also show
that a scale-invariant statistic cannot distinguish translation from
attenuation of a transferred decision variable, although the two call for
opposite remedies. We close with reporting recommendations that cost one
forward pass. Code, data, and a one-cell reproduction are
released.\footnote{\url{https://github.com/pvarshh/quantinterp}}
\end{abstract}


\section{Introduction}
\label{sec:intro}

This paper is about measurement validity. A similarity between two
\emph{estimated} quantities is a random variable whose expectation is set by
the estimator's sampling noise: it has a floor, the value two estimates of the
\emph{same} underlying quantity reach by chance agreement of their errors. A
cosine of $0.996$ is evidence of preservation only if it sits meaningfully
above that floor. Whether it does depends on the sample size and geometry
behind the estimate, quantities a reader must be given to interpret the number
at all. Compression-transfer studies report the cosine and omit the rest. Our
claim is not that their conclusions are wrong. It is that, as reported, they
are unevaluable, and that supplying the missing quantity changes what
published numbers can be read to say.

The case study is interpretability under quantization, where the practice is
load-bearing for safety. In the \textbf{read} mode a probe monitors a deployed
model; in the \textbf{control} mode a steering vector or feature clamp changes
its behavior. Both are computed from particular weights, and deployment
usually does not use those weights: 4-bit serving is standard. The
read-mode evidence is contested (Sec.~\ref{sec:related}). For the control
mode, three papers flag the gap rather than fill it, and the one study
reporting that the refusal direction survives quantization, at cosine
$0.996$~\citep{chhabra2025refusal}, reports no quantized-model attack-success
rate and no $n$ for the cosine. Before running the missing experiment, we ask
a prior question: \emph{would we be able to tell?}

\paragraph{Contributions}
\begin{enumerate}\itemsep1pt
\item A closed-form noise floor for difference-in-means direction similarity,
      governed by $\kap = n\rho^2/d$. It is the classical
      correction-for-attenuation
      argument~\citep{spearman1904proof,spearman1910correlation,brown1910experimental}
      in vector form, validated against Monte Carlo to $0.0015$
      (Sec.~\ref{sec:floor}).
\item \textbf{A measurement of $\rho$ on real activations}
      (Sec.~\ref{sec:measured}), the first we are aware of in the
      compression-transfer literature in a form (with $n$ and $d$) that lets a
      published cosine be placed against its own floor.
\item The correct null for a cross-precision cosine: the split-half null
      measured \emph{within} each quantized model, which separates a rotated
      direction from a merely noisier estimator (Sec.~\ref{sec:application}).
\item Translation and subspace attenuation have opposite signatures and
      remedies, and a scale-invariant statistic distinguishes neither
      (Sec.~\ref{sec:shiftgain}).
\end{enumerate}

\section{Related work}
\label{sec:related}

The two systematic read-mode studies we know of measure different axes.
\citet{duan2026monitors} finds quantization nearly inert for probe
\emph{rankings} (median $\Delta$AUC $-0.0021$ to $-0.0062$ across 840
quantization cells), with fine-tuning the real hazard.
\citet{kadadekar2026quality} finds refusal falling 12--68 percentage points
under INT4 while output quality holds, and probes weak as detectors of the
damage. A stable probe and a degraded behavior can
coexist.\footnote{Both are single-author preprints from the University of
Michigan, this paper's institution; one supplies our framing premise and the
other our motivating quotation. We had no role in either.} At the feature level,
\citet{duan2026perplexity} shows INT7 can \emph{improve} perplexity while
degrading 18.7\% of active SAE features, an aggregate metric missing artifact
damage, and \citet{gupte2025transferability} test causal SAE transfer under
pruning only. Of these quantities, only intervention efficacy governs
control. Quantization error is input-dependent and dominated by
outlier channels~\citep{chang2025inputs,dettmers2022llmint8,sun2024massive}.
Closest to our measurement, \citet{braun2025steering} measure a $d'$-style
separation \emph{along} a fixed steering direction on real activations;
contribution 2 therefore claims only the full-dimensional $\rho$ reported
together with $n$ and $d$ in the compression-transfer setting. Finally, an
existing critique literature argues cosine similarity misleads through
embedding \emph{geometry}~\citep{you2025semantics,nielsen2026cosine}. Our
claim is orthogonal and survives perfect geometry: even for isotropic Gaussian
activations, a cosine between estimated directions is uninterpretable without
the estimator's floor.

\section{The cosine floor}
\label{sec:floor}

The result of this section in one sentence: two independent runs of the
difference-in-means estimator agree, in expectation, at $(1+4/\kap)^{-1}$, a
function of $\kap = n\rho^2/d$ and of nothing else, and simulation confirms
the formula to $0.0015$ across three orders of magnitude in $\kap$.

Let harmful activations be $h^+ \sim \mathcal{N}(\mu_p, \sigma^2 I_d)$ and
harmless $h^- \sim \mathcal{N}(\mu_n, \sigma^2 I_d)$, $n$ samples per class.
The difference-in-means estimator $\hat r = \overline{h^+} - \overline{h^-}$
estimates $r = \mu_p - \mu_n$ with $\|r\| = \delta$; write
$\rho = \delta/\sigma$. Since $\mathrm{Var}(\hat r) = (2\sigma^2/n) I_d$, two
estimates from disjoint halves carry noise variance $4\sigma^2/n$ each, giving
$\mathbb{E}[\hat r_a \!\cdot\! \hat r_b] = \delta^2$ and
$\mathbb{E}[\|\hat r_a\|^2] = \delta^2 + 4d\sigma^2/n$, hence
\begin{equation}
\mathbb{E}\big[\cos(\hat r_a, \hat r_b)\big] \;\approx\;
\frac{1}{1 + 4/\kap}, \qquad \kap \;\equiv\; \frac{n\rho^2}{d}.
\label{eq:floor}
\end{equation}
Equation~\eqref{eq:floor} is the vector form of the classical
correction-for-attenuation, or split-half reliability,
argument~\citep{spearman1904proof,spearman1910correlation,brown1910experimental}:
a similarity between two noisy estimates is attenuated by the reliability of
each. What the compression literature is missing is not the formula but
$\rho$. With it, Eq.~\eqref{eq:floor} is a one-line calculation; without it,
no reported cosine can be placed.

We check the approximation by direct simulation over eleven settings spanning
$\kap \in [0.77, 627]$ and four hidden sizes, including this paper's own two
operating points (Appendix~\ref{app:validation}, Table~\ref{tab:validation}).
Maximum absolute error is $0.0015$; at our operating points it is at most
$0.0001$. Settings with matching $\kap$ but very different $(n, d, \rho)$
return matching floors, so $\kap$, not $\rho$, is the invariant.

\section{Measuring $\rho$}
\label{sec:measured}

Measured on real activations, $\rho$ is $33$--$61$. Per dimension, that is
$54$--$100\times$ above the order-unity regime in which floor formulas are
conventionally exercised, and it is large enough to put the split-half floor
of our own estimator at $0.978$--$0.994$. We measure it on
Qwen2.5-1.5B-Instruct ($d=1536$) with AdvBench harmful
prompts~\citep{zou2023universal} against Alpaca harmless
instructions~\citep{taori2023alpaca}, $n=256$ per class, at six depths. The
estimator is bias-corrected: the plug-in $\|\hat r\|$ overstates $\delta$,
decisively so at small separations (Appendix~\ref{app:bias}). At our operating
point the correction is nearly inert ($61.43 \to 61.33$).

\input{tables/tab_rho}

Table~\ref{tab:rho} carries two findings. First, \textbf{the closed form
matches the resampled floor}, with a maximum discrepancy of $0.0008$ across
six depths. The agreement is not a test of isotropy. The pooled per-dimension
variance makes $d\hat\sigma^2$ an estimate of $\mathrm{tr}\,\Sigma$ for
arbitrary within-class covariance, so anisotropy cancels at leading order; the
column bounds the ratio-of-expectations remainder and any departure from
independence across prompts. Second, \textbf{$\rho$ is large},
and it enters $\kap$ quadratically. At the peak (layer 16), $\kap = 627$ and
the split-half floor of our own estimator is $0.9937$: on this model, at this
$n$, two independent runs agree to $0.994$ \emph{by sampling alone}. We state
$\rho$ per dimension ($\rho/\sqrt{d}$) because a norm over $d$ coordinates
divided by a per-dimension $\sigma$ grows like $\sqrt{d}$ at fixed geometry;
$\kap = n(\rho/\sqrt{d})^2$ is the width-invariant form.

\paragraph{Extrapolating to a published setting}
\citet{chhabra2025refusal} use Llama-2-7B ($d=4096$), so placing their $0.996$
requires carrying our $\rho$ from $d=1536$, and two defensible carries are not
the same claim. \textbf{(A)} holds $\rho$ constant, which implies the wider
model separates the classes less well per dimension. \textbf{(B)} holds the
per-dimension effect constant. Table~\ref{tab:extrapolation} gives both, with
the margin of the reported $0.996$ against each. Under (A) the cosine sits at
or fractionally above its floor from $n=512$ up; under (B) it falls
\emph{below} the floor at $n \geq 512$ and clears it only narrowly at small
$n$. Neither carry is decidable from one hidden size, and $n$ is unreported. Where
we quote one carry it is (A), which weakens our own argument, and the fix is
the same under either: report $\kap$.

\input{tables/tab_extrapolation}

\begin{table}[!t]
\fbox{\parbox{\dimexpr\linewidth-2\fboxsep-2\fboxrule\relax}{%
\textbf{Recommendations.} Each costs at most one forward pass over the
calibration set.
\begin{enumerate}\itemsep1pt
\item \textbf{Report $\kap = n\rho^2/d$}, or at minimum $n$ and $d$, beside
      every similarity statistic between estimated quantities. $\kap$ is
      invariant to hidden size; $\rho$ alone is not.
\item \textbf{Estimate the null inside the compressed model}, not at full
      precision. A full-precision null assumes both estimators have the same
      variance, which is the assumption the null exists to test.
\item \textbf{Report ``no detected movement,'' not ``preservation,''} unless
      an equivalence test (TOST) against a pre-registered margin is run.
\item \textbf{Pair every scale-invariant statistic with a scale-sensitive
      one}: regress compressed decision values on full-precision ones, so
      that translation and attenuation, which call for opposite remedies, can
      be told apart before recalibrating.
\end{enumerate}}}
\label{box:recommendations}
\end{table}

\section{Testing the refusal direction against its own null}
\label{sec:application}

At INT4 the direction rotated; at INT8 no movement is detected
(Table~\ref{tab:rotation}). We extract the difference-in-means refusal
direction of \citet{arditi2024refusal} from full-precision
Qwen2.5-1.5B-Instruct on a fit split of $n=89$ prompts per class, recompute it
on simulated group-wise asymmetric quantized copies, and compare: $0.9999$ at
INT8, $0.9647$ at INT4, against a split-half floor of $0.9820$, which is
Eq.~\eqref{eq:floor} at that $n$ with the layer-16 $\rho$ of
Table~\ref{tab:rho}. The INT4 deficit exceeds twice the spread of the
split-half null measured \emph{within the INT4 model itself}, a deficit that
model's own estimator does not produce. The rest of this section
argues that the within-model null is the right comparison, and that the
obvious alternative errs in the direction that flatters this section.

\paragraph{The null this comparison needs is not the full-precision one}
The tempting argument runs: a cross-precision cosine compares two estimates
from the \emph{same} prompts through near-identical models, whose noise is
shared, so the split-half floor is conservative and anything below it is a
rotation. That argument omits a term the same size as the effect it certifies.
Write $h_q(x) = h_{\mathrm{fp}}(x) + D(x)$ for the quantized activation. The
prompt-constant part of $D$ cancels in a difference of class means, and that
is all the shared-noise argument sees. Its prompt-to-prompt \emph{variation}
does not cancel. It is estimation error present only in the quantized model,
so the quantized estimator has strictly larger variance, and the true null
sits strictly between $1$ and the split-half floor, at a position set by the
size of $D$. Nor can the INT8 row locate it: its weight error is $17\times$
smaller than INT4's, so $0.9999$ confirms the shared-data null at a scale
$17\times$ below the one where the null is used.

\paragraph{Each bit-width against its own null}
The reference a low-bit cosine needs is the split-half cosine of the same
estimator measured \emph{within that quantized model}: disjoint halves of the
fit set at the same $n$, measured through the perturbation, so it absorbs
whatever extra variance the perturbation contributes.
Table~\ref{tab:rotation} reports it per bit-width (``own null'') beside the
identical statistic within FP16 on the same partitions (``FP16 null''), so the
difference between the two columns is a change in the activations, not a
difference of draw. The sign of that difference is the reading: near zero,
with the cosine below both nulls, it is a deficit the estimator does not
produce, while strongly negative means the quantized model is simply the
noisier estimator and accounts for the same deficit with no rotation.

\input{tables/tab_rotation}

\paragraph{The margin is conservative, and the conservatism has a known sign}
The verdict is the decision rule $\cos < \text{own null} - 2\,\mathrm{sd}$,
carried out in Table~\ref{tab:rotation} by the run that measures all three
quantities, with a caption generated to report any outcome. It is a
conservative decision rule, not a
calibrated test, and the conservatism runs in one direction. Under the
no-rotation null, full-precision sampling noise is shared between the two
directions of the paired cross-precision cosine and largely cancels, while
the split-half null carries that noise in full. The own null therefore sits
below the true null, and the rule fires only for deficits the estimator's
self-disagreement cannot produce. The margin is twice the spread over splits,
not the smaller standard error of their mean, because the splits share data
and are not independent draws. A calibrated test would need the distribution
of the paired cosine under the no-rotation null, and that requires a model of
how quantization noise enters the estimate, which is what we do not have. The
conservative rule needs no such model. Its price is resolution, and the
caption states the smallest deficit it can call.

\section{Translation and attenuation are not the same failure}
\label{sec:shiftgain}

AUROC is exactly blind to one of the two ways a transferred decision variable
degrades, and it moves under the other without identifying it. A stable AUROC
therefore does not certify a healthy monitor, and a moving one does not say
what broke. The two failures are a \emph{translation} of the score
distribution and \emph{attenuation} of the signal relative to orthogonal
noise; asymmetric quantizers introduce per-channel offsets, so translation is
live. We construct both with known ground truth ($n=1200$, $d=64$;
Appendix~\ref{app:shiftgain}, Table~\ref{tab:shiftgain}). Under pure
translation, $\Delta$AUROC is $0.0000$ to machine precision, because
translation is rank-preserving, while fixed-threshold accuracy falls from
$0.735$ to $0.550$. A monitor evaluated by AUROC alone is certified healthy
while operating near chance. Attenuation lowers effective SNR and \emph{is}
visible to AUROC, moving it by up to $0.13$. The remedies split the same way:
refitting only the decision threshold recovers $91\%$ of the accuracy lost to
a large translation but $17$--$37\%$ of that lost to attenuation. ``Just
recalibrate'' is sound for one failure mode and close to useless for the
other, and telling them apart requires regressing quantized decision values on
full-precision ones, which no compression-interpretability paper we know of
does. The boxed recommendations collect the fixes; none requires new theory.

\section{Discussion and limitations}
\label{sec:discussion}

The experiment these measurements are prerequisites for, whether an
FP16-derived operator still works on the quantized model, has been run:
$45{,}000$ scored completions. Its rates are withheld because the outcome
measure is a substring refusal classifier with unmeasured agreement against
human judgment; a few-point effect from an instrument of unknown bias is not a
finding. Design, bias-robust observations, and the validation gate that would
release the rates are in Appendix~\ref{app:causal}.

$\rho$ is measured on one model, one hidden size, one dataset;
Sec.~\ref{sec:measured}'s extrapolation is explicitly that, on the carry worse
for our argument. Quantization is simulated and weight-only;
Sec.~\ref{sec:shiftgain}'s perturbations are constructed, so they establish
what estimators can detect, not how often failures occur. One limitation is
also a result about method: a plausible aggregate concealed a data defect, and
a mechanism we had ``confirmed'' rested on an average over conditions that
behave nothing alike (Appendix~\ref{app:audit}).

\begingroup
\small
\bibliographystyle{plainnat}
\bibliography{refs}
\endgroup

\appendix

\section{Validation of the floor formula}
\label{app:validation}

\begin{figure}[h]
\centering
\includegraphics[width=\textwidth]{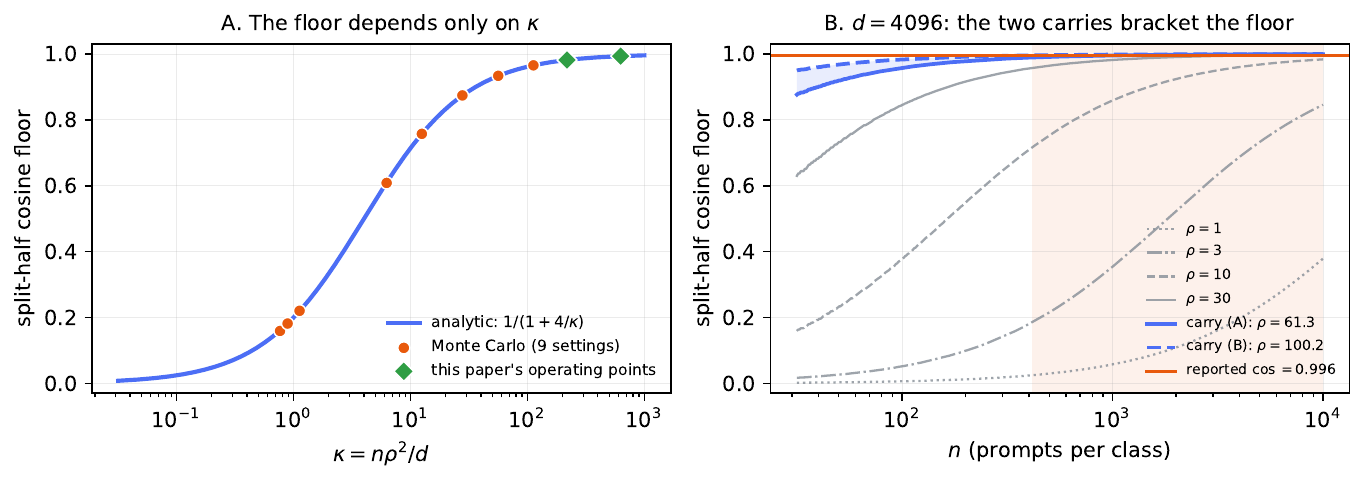}
\caption{\textbf{(A)} Eleven Monte Carlo settings spanning $\kap\in[0.77,627]$
collapse onto the single analytic curve; the diamonds are this paper's two
operating points at the measured peak $\rho$. \textbf{(B)} At $d=4096$ the two
carries of Sec.~\ref{sec:measured} bracket the floor (shaded band); the light
curves show the sweep an unmeasured $\rho$ leaves open, and the reported
$\cos = 0.996$ is marked. In the shaded right region carry (B) puts the floor
above $0.996$, where the reported cosine is not evidence of preservation. Its
abscissa $n$ is still unreported.}
\label{fig:floor}
\end{figure}

Table~\ref{tab:validation} reports the eleven-setting simulation grid behind
Sec.~\ref{sec:floor}. Figure~\ref{fig:floor} shows the settings collapsing
onto the analytic curve, and the two carries of Sec.~\ref{sec:measured}
bracketing the floor at $d=4096$.

\begin{table}[h]
\centering\small
\begin{tabular}{@{}rrrrrrr@{}}
\toprule
$n$ & $d$ & $\rho$ & $\kap$ & Eq.~\eqref{eq:floor} & Monte Carlo & $|$err$|$ \\
\midrule
 400 &   64 &  3.00 & 56.25 & 0.9336 & $0.9339 \pm 0.0120$ & 0.0003 \\
 400 &   64 &  1.00 &  6.25 & 0.6098 & $0.6086 \pm 0.0695$ & 0.0011 \\
 400 &   64 &  0.35 &  0.77 & 0.1607 & $0.1591 \pm 0.1221$ & 0.0015 \\
 512 & 4096 &  3.00 &  1.12 & 0.2195 & $0.2203 \pm 0.0139$ & 0.0008 \\
 512 & 4096 & 10.00 & 12.50 & 0.7576 & $0.7579 \pm 0.0051$ & 0.0003 \\
 512 & 4096 & 30.00 & 112.5 & 0.9657 & $0.9657 \pm 0.0007$ & 0.0000 \\
2048 & 4096 &  5.00 & 12.50 & 0.7576 & $0.7577 \pm 0.0053$ & 0.0001 \\
 128 & 2304 &  4.00 &  0.89 & 0.1818 & $0.1817 \pm 0.0203$ & 0.0002 \\
1000 & 2304 &  8.00 & 27.78 & 0.8741 & $0.8744 \pm 0.0040$ & 0.0002 \\
  89 & 1536 & 61.33 & 217.9 & 0.9820 & $0.9819 \pm 0.0006$ & 0.0001 \\
 256 & 1536 & 61.33 & 626.9 & 0.9937 & $0.9937 \pm 0.0002$ & 0.0000 \\
\bottomrule
\end{tabular}
\caption{Equation~\eqref{eq:floor} against direct simulation, 300 replicates
per row. Settings with matching $\kap$ but very different $(n,d,\rho)$ give
matching floors, as predicted. The last two rows are this paper's own
operating points (Secs.~\ref{sec:measured} and~\ref{sec:application}) at the
measured peak $\rho$. Maximum absolute error $0.0015$, at $\kap = 0.77$.}
\label{tab:validation}
\end{table}

\section{The bias-corrected $\rho$ estimator}
\label{app:bias}

The plug-in $\|\hat r\|$ overstates $\delta$ because the estimator's own noise
contributes $2d\sigma^2/n$ to the squared norm. We subtract it:
$\hat\delta^2 = \|\hat r\|^2 - d\sigma^2(1/n_+ + 1/n_-)$. The correction
matters in high dimension. Against known ground truth at $d=2048$ with $n=256$
per class, the naive estimator returns $\rho = 5.00$ when the truth is $3.0$
and $4.03$ when the truth is $0.5$; the corrected estimator returns $2.99$
(se $0.007$) and $0.44$ (se $0.022$), over 300 replicates. The naive values
are nearly equal across a six-fold difference in true separation because at
this $n$ and $d$ the plug-in norm is dominated by its own noise ($2d/n = 16$
in $\rho^2$ units) and carries almost no signal. The corrected estimator is
not unbiased either. $\hat\delta^2$ is clamped at zero when the subtraction
goes negative, which happens in $31\%$ of replicates at $\rho = 0.5$, so the
estimate there is biased low and its standard error, not its second decimal,
is the resolution. That regime is far from ours: at the measured separations
$\|\hat r\|^2$ exceeds the noise term by two orders of magnitude and the clamp
never binds. A reader carrying this correction to a weakly separated pair
should check the clamp rate, which the released
\texttt{measure\_rho.py -{}-validate} reports. Uncorrected values inflate the
floor in exactly the direction that flatters this paper's argument; at our
operating point the correction moves the floor by under $2\times10^{-4}$.

\section{Translation vs.\ attenuation: full grid}
\label{app:shiftgain}

\begin{table}[h]
\centering\small
\begin{tabular}{@{}lccccccc@{}}
\toprule
& \multicolumn{2}{c}{decision-value fit} & & \multicolumn{2}{c}{AUROC} &
\multicolumn{2}{c}{accuracy} \\
\cmidrule(lr){2-3}\cmidrule(lr){5-6}\cmidrule(l){7-8}
perturbation & slope & $R^2$ & & FP & transfer & transfer & recal. \\
\midrule
none                       & 1.000 & 1.0000 & & 0.8037 & 0.8037 & 0.7354 & 0.7188 \\
translation (small)        & 1.000 & 1.0000 & & 0.8037 & 0.8037 & 0.7229 & 0.7188 \\
translation (large)        & 1.000 & 1.0000 & & 0.8037 & 0.8037 & 0.5500 & 0.7188 \\
attenuation ($0.6\times$)  & 0.686 & 0.9472 & & 0.8037 & 0.7591 & 0.6750 & 0.6854 \\
attenuation ($0.3\times$)  & 0.451 & 0.7168 & & 0.8037 & 0.6702 & 0.5792 & 0.6354 \\
amplification ($1.5\times$)& 1.392 & 0.9793 & & 0.8037 & 0.8251 & 0.7438 & 0.7188 \\
translation $+$ attenuation& 0.686 & 0.9472 & & 0.8037 & 0.7591 & 0.6479 & 0.6854 \\
\bottomrule
\end{tabular}
\caption{Known perturbations along the probe direction, $n=1200$, $d=64$.
Translation leaves AUROC \emph{exactly} unchanged while calibrated-threshold
accuracy falls to $0.55$; attenuation moves AUROC by up to $0.13$. The slope
is the regression of quantized on full-precision \emph{decision values}, not
the injected subspace gain $g$: the decision variable mixes the perturbed
direction with unperturbed ones, so the slope is $1 - c(1-g)$ with $c$ the
share of decision-variable variance the perturbed direction carries
($c = 0.784$ here, consistent across all three gain rows). It therefore
\emph{understates} subspace attenuation (slope $0.686$ for an injected
$g = 0.6$) and detects it rather than measures it; recovering $g$ itself
requires projecting onto the probe direction before regressing.}
\label{tab:shiftgain}
\end{table}

Refitting only the decision threshold recovers $91.0\%$ of the accuracy lost
to a large translation, but $17.2\%$ and $37.0\%$ of that lost to attenuation,
and $42.9\%$ under combined translation and attenuation (reported only where
the accuracy gap exceeds $0.05$; on the two smaller-gap rows the ratio has a
near-zero denominator and is not meaningful). The denominator throughout is
the gap between transfer and a probe \emph{retrained} at that perturbation,
not the gap to full-precision accuracy. Those coincide for three of the four
rows but not for the severe-attenuation case, so recomputing these from
Table~\ref{tab:shiftgain}'s columns alone gives $36.0\%$ rather than $37.0\%$
there.

\section{The deferred causal-transfer arm}
\label{app:causal}

\paragraph{Design}
$n=250$ prompts per condition (harmless prompts for the additive operators,
harmful for \texttt{ablate}), five fit/eval splits, three bit-widths, four
operators and three coefficients: $45{,}000$ scored completions. The
difference-in-means direction every operator is built from, and directional
ablation itself, are \citet{arditi2024refusal}'s; what is tested is whether
they survive a change of precision. The direction is fit on a prompt set
disjoint from the evaluation set, without which ablation is circular. Effects
are headroom-normalized as $(\text{rate}-\text{base})/(1-\text{base})$ with
the sign flipped for suppression, because quantization moves the baseline and
raw differences are not comparable across bit-widths. The five seeds resample
the fit/eval split from the same $520$ AdvBench prompts, so they measure split
variance, not independent replication.

\paragraph{Why the rates are withheld}
All $45{,}000$ completions are labelled by substring matching over refusal
prefixes, and its agreement with human judgement is unmeasured. The $200$ hand
labels that would establish it do not exist, so no agreement coefficient
against human ground truth can be quoted. The additive arm is where that
matters most: its effects are a few points over a low baseline, inside the
range a classifier bias of unknown size can manufacture. The inferential
apparatus is incomplete in the same direction. The available contrasts are
unpooled two-proportion $z$ tests on rates aggregated over seeds, while the
paired McNemar and DeLong tests the design was sized for need per-example
scores joined across bit-widths, and the TOST any \emph{preservation} claim
would require is not computed. A paper whose argument is that a statistic must
be reported alongside the quantity that makes it interpretable should not
quote a rate a reader cannot place.

\paragraph{What survives the caveat}
Two observations are robust to an uncalibrated instrument, and we state both
qualitatively. First, \texttt{add\_relative}, which scales the injected vector
by $\|h\|$, has a usable range one coefficient wide. At $c{=}1$ it is the
strongest operator we test, driving refusal on harmless prompts from near
baseline to near ceiling; at $c{=}4$ it reaches exactly zero in every
bit-width and every seed, while \texttt{add\_absolute} over the same sweep is
monotone with no cliff. A rate near ceiling and a rate of exactly zero are
further apart than a classifier bias can manufacture; the few-point additive
differences are not. The proposed mechanism, $\|h\|$ dominated by outlier
channels orthogonal to the steered subspace~\citep{sun2024massive}, is
consistent but unshown; since output quality was not measured, a rate of zero
is also consistent with degenerate generation. Second, directional ablation,
which is scale-free, suppresses refusal on harmful prompts by a large fraction
of available headroom at every bit-width tested. Neither observation licenses
a bit-width comparison; whether the direction moved at INT4 is settled in
Sec.~\ref{sec:application} by a measurement that does not depend on this arm.

\paragraph{What would close it}
(i) The $200$ blind hand labels and a weighted agreement coefficient against
them; this gates the rest. (ii) Per-example scores joined across bit-widths,
so paired McNemar and DeLong tests replace aggregated $z$ tests. (iii) A
pre-registered equivalence margin and a TOST, without which nothing in this
arm can support a preservation claim. The code for all three and the completed
run are in the repository; the labels are the only human-time item.

\section{Self-audit: errors made and what caught them}
\label{app:audit}

The most serious limitation of this work is a data defect diagnosed after the
pilot, and it is instructive. The harmless-prompt refusal baseline was
$0.375$, which we initially attributed to a substring classifier firing on
hedged completions. Inspecting the flagged generations showed otherwise: $23$
of $24$ were triggered by \texttt{"i'm sorry"} on responses such as \emph{``I'm
sorry, but you haven't provided any passage for me to summarize.''} Alpaca
rows carry an \texttt{input} field holding the passage the instruction
operates on; loading \texttt{instruction} alone yields prompts like
``Summarize the given passage.'' with no passage. The model's response is
correct, and the classifier's label is arguably correct too. \emph{The prompts
were malformed.} The harmful side was clean: $64/64$ flagged, every one an
unambiguous refusal. We repaired the loader and re-ran both affected
measurements; the harmless baseline fell from $0.375$ to $0.05$. The counts
come from the pilot's inspection pass, whose artifacts are unreleased for the
reason in Appendix~\ref{app:causal}. What is checkable in the repository is
the repair, pinned by a regression test that fails if the
\texttt{input}-column filter is dropped.

Two predictions then failed, and the failures are the paper's thesis turned on
itself. First, we had argued that a heterogeneous harmless class inflates
within-class variance and therefore \emph{depresses} $\rho$, making the
measured values a lower bound. Measured, the contamination did not act in one
direction: $\rho$ \emph{fell} at three early layers ($40.9\to34.3$,
$38.8\to33.6$, $36.3\to32.9$) and \emph{rose} at three late ones
($60.1\to61.3$, $54.9\to59.0$, $52.5\to56.1$). There was no lower bound to
have. We had flagged the claim as unverified rather than asserting it, and it
was still wrong. That is the argument for measuring, not for hedging more
precisely. Second, on the contaminated prompts \texttt{add\_relative} was
negative at every bit-width, which we read as over-injection driven by
$\|h\|$. The mechanism is real; the evidence was not. That number was a mean
over coefficients $c \in \{1,2,4\}$, whose three behaviors are nothing alike:
strongest operator at $c{=}1$, exactly zero at $c{=}4$. An average over
conditions that behave nothing alike is a different quantity, not a summary of
them, and it happened to have the sign of the mechanism we then explained. A
scale-invariant statistic is not the only thing that can be uninformative
while looking decisive. So can an average, and so can an explanation that fits
it.

\section{Ethics and reproducibility}
\label{app:ethics}

This paper measures whether interpretability artifacts survive quantization;
it does not introduce a capability. The one intervention that suppresses
refusal, directional ablation, is the published method of
\citet{arditi2024refusal}, applied to open-weight models whose refusal
behavior that method, fine-tuning, or system prompts can already remove. Our
contribution is the measurement of whether the FP16-derived direction still
works after quantization. All prompts come from public benchmark datasets
(AdvBench, Alpaca); no new harmful prompts were written and no jailbreaks are
published. The released result files contain summary statistics over
activations and no generations; the steering run's completions are withheld
for the methodological reason in Appendix~\ref{app:causal}.

All results reproduce from a single self-contained notebook in the
repository.\footnote{\url{https://github.com/pvarshh/quantinterp}}
Tables~\ref{tab:validation} and~\ref{tab:shiftgain} need neither a GPU nor
any dataset and are regenerated by CI on every push. Tables~\ref{tab:rho}
and~\ref{tab:extrapolation} come off one \texttt{measure\_rho} run, about
three minutes on a free Colab T4; Table~\ref{tab:rotation}, including the
within-model nulls, is one further cell in the same session. Every run writes
a manifest recording git commit, package versions, seed, and resolved dataset
source. The two measurement artifacts behind
Secs.~\ref{sec:measured}--\ref{sec:application} are committed with their
manifests, so every generated table rebuilds from the repository with no GPU
and no gated dataset. AdvBench's Hugging Face mirror is gated, a separate
permission from the model; Alpaca is not.

\end{document}

%% file: tables/tab_rho.tex
\begin{table}[t]
\centering\small
\begin{tabular}{@{}rrrrrrrr@{}}
\toprule
layer & $\rho$ (naive) & $\rho$ & $\rho/\sqrt{d}$ & $\kap$ & floor, Eq.~\eqref{eq:floor} & floor, measured & $\Delta$ \\
\midrule
4 & 34.47 & 34.29 & 0.875 & 196.0 & 0.9800 & 0.9808 & 0.0008 \\
8 & 33.72 & 33.54 & 0.856 & 187.5 & 0.9791 & 0.9799 & 0.0008 \\
12 & 33.05 & 32.87 & 0.839 & 180.1 & 0.9783 & 0.9790 & 0.0008 \\
16 & 61.43 & \textbf{61.33} & \textbf{1.565} & 627.0 & 0.9937 & \textbf{0.9939} & 0.0002 \\
20 & 59.12 & 59.02 & 1.506 & 580.6 & 0.9932 & 0.9935 & 0.0003 \\
24 & 56.22 & 56.11 & 1.432 & 524.8 & 0.9924 & 0.9928 & 0.0004 \\
\bottomrule
\end{tabular}
\caption{\textbf{Measured $\rho$ on real activations.} Qwen/Qwen2.5-1.5B-Instruct, $d=1536$, $n=256$ per class, walledai/AdvBench+tatsu-lab/alpaca. Separation is $33$--$61$ in $\rho$ units: $0.84$--$1.56$ pooled within-class SDs \emph{per dimension}, the column comparable across hidden sizes ($\kap = n\rho^2/d = n(\rho/\sqrt{d})^2$; Sec.~\ref{sec:measured}). An order-unity $\rho$ (the natural guess absent any reported measurement) is $0.016$ SDs per dimension at $d=4096$, $54$--$100\times$ below the range measured here. The closed form predicts the empirically measured split-half floor to within $0.0008$ at every depth. Bold marks the peak, layer 16.}
\label{tab:rho}
\end{table}

%% file: tables/tab_extrapolation.tex
\begin{table}[t]
\centering\small
\begin{tabular}{@{}rcccc@{}}
\toprule
$n$ & (A) floor & (A) margin & (B) floor & (B) margin \\
\midrule
100 & 0.9583 & $+0.0377$ & 0.9839 & $+0.0121$ \\
256 & 0.9833 & $+0.0127$ & 0.9937 & $+0.0023$ \\
512 & 0.9916 & $+0.0044$ & 0.9968 & $-0.0008$ \\
1024 & 0.9958 & $+0.0002$ & 0.9984 & $-0.0024$ \\
\bottomrule
\end{tabular}
\caption{\textbf{Carrying the measured $\rho$ to $d=4096$.} ``margin'' is $0.996 - \text{floor}$: positive means the published cosine clears its own noise floor. Peak $\rho = 61.33$ at $d=1536$ (layer 16, Table~\ref{tab:rho}). \textbf{(A)} holds $\rho$ constant; \textbf{(B)} holds the per-dimension effect $1.565$ constant, giving $\rho = 100.16$. Under carry (B) the reported $\cos = 0.996$ is at or below its own floor for $n \geq 512$, so it is not evidence of preservation there.}
\label{tab:extrapolation}
\end{table}

%% file: tables/tab_rotation.tex
\begin{table}[t]
\centering\small
\setlength{\tabcolsep}{4pt}
\begin{tabular}{@{}lcccccccl@{}}
\toprule
 & rel.\ wt.\ err. & subspace gain & $\cos$ to FP16 & own null & sd & FP16 null & floor, Eq.~\eqref{eq:floor} & verdict \\
\midrule
INT8 & 0.0063 & 1.007 & 0.9999 & 0.9827 & 0.0047 & 0.9825 & 0.9820 & within noise \\
INT4 & 0.1066 & 0.933 & \textbf{0.9647} & 0.9803 & 0.0052 & 0.9825 & 0.9820 & \textbf{rotation} \\
\bottomrule
\end{tabular}
\caption{\textbf{Rotation, against each bit-width's own null.} At INT4 the cosine falls below the split-half null measured within that model itself by more than twice the null's own split-to-split spread ($0.9647$ vs.\ $0.9803 - 2(0.0052)$), so the deficit exceeds that bit-width's own estimation noise by a margin the noise does not produce, and the rotation is a detection. The verdict is the decision rule $\cos < \text{own null} - 2\,\mathrm{sd}$: ``own null'' is the split-half cosine measured \emph{within} that quantized model (200 random splits at $n=89$ per class), sd its spread over those splits, and ``FP16 null'' the identical statistic within FP16 on the same partitions (same $n$, same splits, same seed), so the difference between the two null columns (INT8 $+0.0002$; INT4 $-0.0022$) is the change in the activations and not the draw. It is a conservative decision rule, not a calibrated test; Sec.~\ref{sec:application} argues why, and why the conservatism has a known sign. The smallest deficit this rule can call is $2\,\mathrm{sd} \leq 0.0104$ in cosine units; a rotation smaller than that is invisible at this $n$, so ``within noise'' bounds the detectable effect and does not assert absence. The floor of Eq.~\eqref{eq:floor}, $0.9820$, is computed from Table~\ref{tab:rho} at layer 16 rather than stored beside the cosine, and decides no row; its distance from the FP16 null, $+0.0006$, is Sec.~\ref{sec:measured}'s isotropic-Gaussian check repeated at this $n$. Subspace gain is $\overline{|h_q\!\cdot\!\vhat|}/\overline{|h_\mathrm{fp}\!\cdot\!\vhat|}$.}
\label{tab:rotation}
\end{table}